\documentclass[runningheads]{llncs}
\usepackage{graphicx}
\usepackage{tabularx}
\usepackage{multirow}
\usepackage[depth=2]{bookmark}
\usepackage[utf8]{inputenc}
\usepackage[T1,T2A]{fontenc}
\usepackage[russian, english]{babel}
\usepackage{csquotes}
\usepackage[space]{cite}
\usepackage{float}
\usepackage{xcolor}
\usepackage{hyperref}
\usepackage{amsmath}
\hypersetup{pdfstartview=FitH,  linkcolor=blue,urlcolor=blue, colorlinks=true}
\begin{document}
	\title{Lexicon-based Methods vs. BERT\\for Text Sentiment Analysis}

	\author{Anastasia Kotelnikova\inst{1}\orcidID{0000-0001-9942-680X} \and
	Danil Paschenko\inst{1}\orcidID{0000-0003-2671-8208} \and
	Klavdiya Bochenina\inst{2}\orcidID{0000-0001-6025-0552} \and
	Evgeny Kotelnikov\inst{1,2}\orcidID{0000-0001-9745-1489}}
\authorrunning{A. Kotelnikova et al.}
\institute{Vyatka State University, Kirov, Russia\\ \email{kotelnikova.av@gmail.com}\and
ITMO University, Saint-Petersburg, Russia
}

	\maketitle              % typeset the header of the contribution
	\begin{abstract}The performance of sentiment analysis methods has greatly increased in recent years. This is due to the use of various models based on the Transformer architecture, in particular BERT. However, deep neural network models are difficult to train and poorly interpretable. An alternative approach is rule-based methods using sentiment lexicons. They are fast, require no training, and are well interpreted. But recently, due to the widespread use of deep learning, lexicon-based methods have receded into the background. The purpose of the article is to study the performance of the SO-CAL and SentiStrength lexicon-based methods, adapted for the Russian language. We have tested these methods, as well as the RuBERT neural network model, on 16 text corpora and have analyzed their results. RuBERT outperforms both lexicon-based methods on average, but SO-CAL surpasses RuBERT for four corpora out of 16.
	\end{abstract}
	\setcounter{footnote}{0}
	\section{Introduction}
	The performance of sentiment analysis has improved dramatically\footnote{See for example: https://paperswithcode.com/task/sentiment-analysis.}  over the past few years, for example:
	\begin{itemize}
		\item the English-language corpus SST-5 (Stanford Sentiment Treebank -- 5 classes), the accuracy increased from 45.70 (RNTN model \cite{socher-etal-2013}) to 59.10 ({R}o{BERT}a-large+Self-Explaining \cite{sun-etal-2020});
		\item for the English-language corpus Yelp Reviews (5 classes), the error decreased from 37.95 (Char-level CNN model \cite{zhang-etal-2015}) to 27.05 (XLNet model \cite{Yang-etal-2019});
		\item for the Russian-language news corpus ROMIP-2012 (3 classes) \cite{chetviorkin-loukachevitch-2013} F1-score increased from 62.10 (lexicon-based Polyarnik system) \cite{kuznetsova-etal-2013} to 72.69 \cite{golubev-loukachevitch-2021}.
        \end{itemize}
    This improvement in performance is mainly associated with the development of deep learning methods, especially various models based on the Transformer architecture \cite{vaswani-etal-2017}, in particular, BERT \cite{devlin-etal-2019}.
  
    However, deep neural network models are difficult to train: they need large amounts of data; training requires powerful expensive video cards with a large memory size; the learning process is time consuming and energy intensive \cite{li-2018}. Another issue is the complexity of interpreting the results of the models \cite{belinkov-etal-2020}.
  
    An alternative are rule-based (or lexicon-based) methods using sentiment lexicons \cite{taboada-2016}. They are fast, do not require training and are well interpreted \cite{birjali-etal-2021}. But recently, due to the widespread use of deep learning, lexicon-based methods have receded into the background.
    
    For the Russian language, there are several recent studies of deep learning models for sentiment analysis \cite{smetanin-komarov-2021, golubev-loukachevitch-2021}. However, there are currently no studies devoted to comparing lexicon-based methods and deep learning models.
    
    We strive to close this gap and compare the fine-tuned deep neural network model RuBERT \cite{kuratov-arkhipov-2019} with two lexicon-based methods adapted for the Russian language – SO-CAL \cite{taboada-etal-2011} and SentiStrength \cite{thelwall-etal-2010}. For testing we have used 16 Russian-language text corpora, labelled by sentiment into 3 classes.
    
    The contribution of this article is as follows:

    \begin{itemize}        
        \item lexicon-based methods SO-CAL and SentiStrength have been adapted for the Russian language;
        \item performance evaluation of lexicon-based methods and RuBERT has been carried out for 16 Russian-language text corpora;
        \item the performance of SO-CAL and SentiStrength for 17 sentiment lexicons have been estimated: 9  publicly available Russian sentiment lexicons and  a set of 8 combined lexicons;
        \item the classification results of lexicon-based methods and RuBERT have been analyzed.
	\end{itemize}

	\section{Lexicon-based Methods and Tools}
There are several tools for sentiment analysis on the base of sentiment lexicons:
\begin{itemize} 
        \item open source: SO-CAL \cite{taboada-etal-2011}, VADER \cite{hutto-gilbert-2014}, Pattern, TextBlob;
        \item proprietary: SentiStrength \cite{thelwall-etal-2010}, SentText \cite{schmidt-etal-2021}.
\end{itemize}

Taboada et al. developed SO-CAL\footnote{https://github.com/sfu-discourse-lab/SO-CAL.} (Semantic Orientation CALculator) – a method and tool for determining the sentiment of texts in English and Spanish \cite{taboada-etal-2011}. The sentiment is recognized on the basis of counting the weights of the sentiment words included in the text (only nouns, adjectives, verbs and adverbs are taken into account). A system of rules is also involved to account for the influence of lexical markers such as modifiers, negations, and irrealis markers. \textit{Modifiers} are lexical markers that increase (e.g., \textit{very}, \textit{the most}) or decrease (e.g., \textit{slightly}, \textit{somewhat}) the intensity of the next sentiment word. \textit{Negations} (e.g., \textit{not}, \textit{nothing}) either invert the polarity of the next sentiment word, or shift its intensity towards the opposite polarity (for example, in SO-CAL, a shift is used, and in VADER -- an inversion with a certain coefficient). \textit{Irrealis markers} indicate that the sentiment score for a given sentence should not be taken into account. These markers are modal verbs (e.g., \textit{could}, \textit{should}), conditional words (e.g., \textit{if}), some verbs (e.g., \textit{expect}, \textit{doubt}), a question mark, and quoted words.

Hutto and Gilbert proposed VADER\footnote{https://github.com/cjhutto/vaderSentiment.} (Valence Aware Dictionary for sEntiment Reasoning) – a lexicon, method and tool for sentiment analysis of English texts \cite{hutto-gilbert-2014}. The sentiment lexicon was built from the well-known dictionaries LIWC, ANEW, and General Inquirer and then crowdsourced in sentiment intensity. Also, emoticons, acronyms and slang were included into the lexicon. The VADER takes into account exclamation marks, capitalization, modifiers, negations and contrasts.

Pattern\footnote{https://github.com/clips/pattern.}  is a web mining library that supports sentiment analysis in English and French\cite{desmedt-daelemans-2012}. For this, a lexicon of sentiment adjectives is used, often found in product reviews.

TextBlob\footnote{https://textblob.readthedocs.io.}  is a text processing library that includes two components for sentiment analysis – based on a naive Bayesian classifier and an implementation from the \textit{Pattern} library.

Thelwall et al. developed SentiStrength\footnote{http://sentistrength.wlv.ac.uk.}, a tool for sentiment analysis of short social media texts based on the method bearing the same name\cite{thelwall-etal-2010}. The tool gives two scores for the input text: a negative score from –1 to –5 and a positive score from +1 to +5. The decision is based on a list of sentiment words with weights corresponding to the sentiment intensity. The method also takes into account modifiers, negations, question words, slang, idioms and emoticons.

Schmidt et al. developed SentText\footnote{https://thomasschmidtur.pythonanywhere.com.}, a web-based sentiment analysis tool in the digital humanities\cite{schmidt-etal-2021}. The original version of SentText was developed for the German language using the SentiWS and BAWL-R dictionaries. This tool takes negations into account. SentText has the ability to visualize the results, including highlighting the sentiment words, information about the polarity of individual words and the text as a whole, as well as comparing texts by sentiment.

Of these tools, as far as we know, only SentiStrength has two adaptations for the Russian language\footnote{Given on the website: http://sentistrength.wlv.ac.uk.}, but we could not find a detailed description of their implementations.

In our work, we have adapted SO-CAL and SentiStrength for the Russian language. SO-CAL is the most advanced open source sentiment analysis tools. SentiStrength, despite being proprietary software, makes it easy to adapt to a new language. For this purpose, it is necessary to provide it with a sentiment lexicon in the target language and other linguistic resources: lists of modifiers, negations, interrogative words, slang and idioms.

\section{Materials and Methods}

\subsection{Lexicon-based Methods Adaptation}

The adaptation of SO-CAL and SentiStrength to the Russian language includes the following steps:

\begin{itemize} 
\item morphological analysis of input texts based on RNNMorph\footnote{https://github.com/IlyaGusev/rnnmorph.};
\item preparation of a Russian-language sentiment lexicon. The existing lexicons were used to form it and are described in Subsection~\ref{sentiment_lexicons}. A peculiarity of SO-CAL is to take into account only nouns, adjectives, verbs, adverbs;
\item preparation of Russian-language lists of modifiers (e.g., \textit{очень}, \textit{едва}, \sloppy\textit{значительно}) and negations (e.g., \textit{не}, \textit{без}, \textit{невозможно}). They were obtained by translating the corresponding SO-CAL and SentiStrength lists, as well as by adding Russian-language synonyms;
\item preparation for SO-CAL of Russian-language lists of irrealis markers (e.g., \textit{ожидать}, \textit{можно}, \textit{кто-нибудь});
\item modification of the SO-CAL source code for processing texts with the results of Russian morphological analysis;
\item organization of programming interface with the desktop version of SentiStrength to submit input texts and process its results.
\end{itemize}

\subsection{Sentiment Lexicons} \label{sentiment_lexicons}

The key resource for the considered sentiment analysis methods is the sentiment lexicon. The performance of sentiment analysis depends on the completeness and accuracy of such a lexicon. We have used 9 publicly available Russian sentiment lexicons (two of them – EmoLex and Chen-Skiena’s – are Russian versions of multi-lingual lexicons) \cite{kotelnikov-etal-2018}.

Each lexicon has been processed as follows:

\begin{itemize} 
\item	neutral words have been removed (if such were present in the lexicon);
\item	words that are both positive and negative in the lexicon have been removed (including the analysis of words with the spelling “e” and “ё”);
\item	words containing Latin letters have been removed;
\item	all words have been converted to a lower case;
\item	words have been normalized using RNNMorph;
\item	only one occurrence of each element has been left (an element can be a separate word or phrase).
\end{itemize}
The characteristics of the lexicons are shown in Table~\ref{tab1}.

\begin{table}
  \begin{center}
  \caption{The characteristics of sentiment lexicons.}
  \renewcommand\tabcolsep{5.0pt}
  \label{tab1}  
    \begin{tabular}{p{35mm}p{12mm}p{12mm}p{12mm}p{12mm}p{12mm}}
    \hline
    \hline
    Lexicon & Total & \multicolumn{2}{l}{Positive elements} & \multicolumn{2}{l}{Negative elements}  \\ 
    & &\# & \% & \# & \%  \\ 
    \hline
    RuSentiLex\cite{loukachevitch-levchik-2016} & 12,560 & 3,258 & 25.9\% & 9,302 & 74.1\% \\ 
    Word Map\cite{kulagin-2019} & 11,237 & 4,491 & 40.0\% & 6,746 & 60.0\% \\
    SentiRusColl\cite{kotelnikova-kotelnikov-2019} & 6,538 & 3.981 & 60.9\% & 2,557 & 39.1\% \\ 
    EmoLex\cite{mohammad-turney-2013} & 4,600 & 1,982 & 43.1\% & 2,618 & 56.9\% \\
    LinisCrowd\cite{koltsova-etal-2016} & 3,986 & 1,126 & 28.2\% & 2,860 & 71.8\% \\
    Blinov’s lexicon\cite{blinov-etal-2013} & 3,524 & 1,611 & 45.7\% & 1,913 & 54.3\% \\ 
    Kotelnikov’s lexicon\cite{kotelnikov-etal-2016} & 3,206 & 1,028 & 32.1\% & 2,178 & 67.9\% \\ 
    Chen-Skiena’s lexicon\cite{chen-skiena-2014} & 2,604 & 1,139 & 43.7\% & 1,465 & 56.3\% \\ 
    Tutubalina’s lexicon\cite{tutubalina-2016} & 2,442 & 1,032 & 42.3\% & 1,410 & 57.7\% \\		
    \hline   
    \hline
   \end{tabular}
  \end{center}
\end{table}
    
We also used the “voting” procedure of these lexicons to build a set of 8 combined lexicons \textit{Lex1}..\textit{Lex8}: only those words that are included in at least \textit{N} sentiment lexicons are included in the \textit{LexN} lexicon. Thus, \textit{Lex1} includes all sentiment words that occur in at least one lexicon. \textit{Lex9} turned out to be empty – not a single item is included in all sentiment lexicons at the same time. The characteristics of the combined lexicons are shown in  Table~\ref{tab2}.

\subsection{Text Corpora}

For evaluation, we have used 16 public text corpora labelled by sentiment \cite{smetanin-2020} (see Table~\ref{tab3}), including:

\begin{itemize} 
\item	corpora of reviews about books, movies and cameras, as well as news articles from the ROMIP 2011 \cite{chetviorkin-etal-2012} and ROMIP 2012 \cite{chetviorkin-loukachevitch-2013} seminars;
\item	corpora of reviews about cars and restaurants, as well as tweets about banks and telecom companies of the SentiRuEval 2015\cite{loukachevitch-etal-2015}, SentiRuEval 2016\cite{loukachevitch-rubtsova-2016} and SemEval 2016\cite{pontiki-etal-2016} seminars;
\item	the RuSentiment corpus containing posts on VKontakte \cite{rogers-etal-2018};
\item	LinisCrowd corpus, including posts and comments from LiveJournal \cite{koltsova-etal-2016}. We have used texts labelled by one annotator as training data, and texts labelled by several annotators as test data.
\end{itemize}

\begin{table}
  \begin{center}
  \caption{The characteristics of the combined sentiment lexicons.}
  \renewcommand\tabcolsep{5.0pt}
  \label{tab2}    
    \begin{tabular}{p{15mm}p{13mm}p{13mm}p{13mm}p{13mm}p{13mm}}
    \hline
    \hline
    Lexicon & Total & \multicolumn{2}{l}{Positive elements} & \multicolumn{2}{l}{Negative elements}  \\ 
    & &\# & \% & \# & \%  \\ 
    \hline
    Lex1 & 33,080 & 13,443 & 40.6\% & 19,637 & 59.4\% \\ 
    Lex2 & 9,377 & 3,147 & 33.6\% & 6,230 & 66.4\% \\ 
    Lex3 & 4,325 & 1,521 & 35.2\% & 2,804 & 64.8\% \\ 
    Lex4 & 2,313 & 823 & 35.6\% & 1,490 & 64.4\% \\ 
    Lex5 & 1,266 & 475 & 37.5\% & 791 & 62.5\% \\ 
    Lex6 & 607 & 258 & 42.5\% & 349 & 57.5\% \\ 
    Lex7 & 240 & 114 & 47.5\% & 126 & 52.5\% \\ 
    Lex8 & 52 & 31 & 59.6\% & 21 & 40.4\% \\ 
	\hline
	\hline
   \end{tabular}
  \end{center}
\end{table}

\section{Results}

\subsection{Experimental Setup} \label{experimental_setup}

We have tested two lexicon-based methods of sentiment analysis adapted for the Russian language – SO-CAL and SentiStrength, as well as a deep neural network model RuBERT\cite{kuratov-arkhipov-2019}.

In general, lexicon-based methods can be used for sentiment analysis without training. However, since training corpora are available in our experiments, we have used them to tune the hyperparameters of the lexicon-based methods: we have chosen the optimal sentiment lexicons for both methods (out of 17 lexicons – see Subsection~\ref{sentiment_lexicons}) and have determined the thresholds for positive and negative sentiment classes.

\begin{table}
  \begin{center}
  \caption{Corpora characteristics.}
  \renewcommand\tabcolsep{5.0pt}
  \label{tab3}    
    \begin{tabular}{p{25mm}p{26mm}p{8mm}p{9mm}p{10mm}p{10mm}p{10mm}}
    \hline
    \hline
    Corpus & Type & Split & Total &  Positive & Negative & Neutral\\        
\hline
	LinisCrowd & posts & train & 28,853 & 7.7\% & 42.5\% & 49.8\% \\ 
	 &  & test & 14,260 & 9.5\% & 47.3\% & 43.2\% \\ 
	Romip 2011 & book reviews & train & 22,098 & 79.7\% & 9.3\% & 11.0\% \\ 
	 &  & test & 228 & 64.0\% & 6.2\% & 29.8\% \\ 
	 & movie reviews & train & 14,808 & 70.6\% & 12.7\% & 16.7\% \\ 
	 &  & test & 263 & 70.3\% & 10.7\% & 19.0\% \\ 
	 & camera reviews & train & 9,460 & 80.5\% & 10.6\% & 8.9\% \\
	 &  & test & 207 & 61.8\% & 17.9\% & 20.3\% \\ 
	Romip 2012 & book reviews & test & 129 & 77.5\% & 7.0\% & 15.5\% \\
	 & movie reviews & test & 408 & 65.2\% & 15.4\% & 19.4\% \\
	 & camera reviews & test & 411 & 85.4\% & 1.7\% & 12.9\% \\ 
	 & news & train & 4,260 & 26.2\% & 43.7\% & 30.1\% \\ 
	 &  & test & 4,573 & 31.7\% & 41.3\% & 27.0\% \\ 
	SentiRuEval 2015 & car reviews & train & 203 & 56.6\% & 14.8\% & 28.6\% \\
	 &  & test & 200 & 49.0\% & 13.0\% & 38.0\% \\ 
	 & restaurant reviews & train & 200 & 68.0\% & 14.0\% & 18.0\% \\
	 &  & test & 203 & 71.9\% & 12.8\% & 15.3\% \\ 
	 & bank tweets & train & 4,883 & 7.2\% & 21.7\% & 71.1\% \\
	 &  & test & 4,534 & 7.6\% & 14.4\% & 78.0\% \\ 
	 & telecom tweets & train & 4,839 & 18.8\% & 32.7\% & 48.5\% \\ 
	 &  & test & 3,774 & 9.1\% & 22.4\% & 68.5\% \\ 
	SentiRuEval 2016 & bank tweets & test & 3,302 & 9.1\% & 23.1\% & 67.8\% \\
	 & telecom tweets & test & 2,198 & 8.3\% & 45.8\% & 45.9\% \\ 
	SemEval 2016 & restaurant reviews & test & 103 & 67.0\% & 14.6\% & 18.4\% \\
	RuSentiment & posts & train & 24,124 & 38.0\% & 15.2\% & 46.8\% \\
	 &  & test & 2,621 & 36.0\% & 9.8\% & 54.2\% \\ 
	\hline
	\hline
   \end{tabular}
  \end{center}
\end{table}

The thresholds are defined as follows. SO-CAL returns the single sentiment score s for the text to be converted to a class label (positive, negative, or neutral). We have fit two thresholds on the training data – \(t_{pos}\) and \(t_{neg}\). The decision about the sentiment of the text c is made on the basis of the following expression:
\[c =
\begin{cases}
\textit{neutral},\; if\;  s < t_{pos}\; and\; s > t_{neg}, \\
\textit{positive},\; if\; s \geq t_{pos},\\
\textit{negative},\; if\; s \leq t_{neg}. 
\end{cases}\]

SentiStrength returns two sentiment scores for the text – positive \(s_{pos}\)  and negative \(s_{neg}\). We select two coefficients \(k_{neut}\) and \(k\) such that:
\[c =
\begin{cases}
neutral,\; if\; s_{pos}\leq k_{neut}\; and\; s_{neg}\leq k_{neut}, \\
positive,\; if\; s_{pos}>ks_{neg},\\
negative,\; otherwise. 
\end{cases}\]

The results of selecting lexicons and threshold values are given in Subsection~\ref{results_experiments}.

The pretrained RuBERT model was fine-tuned separately on each training corpus with the following hyperparameters: learning rate {$2\cdot10^{-5}$}, number of epochs 5, batch size 12. The results are given on average for five runs to reduce the influence of random weight initialization. The training has been carried out using the Google Colab Pro service on NVIDIA Tesla P100 and V100 video cards.

For all the corpora a three-class problem of sentiment analysis was solved – the classification of texts into positive, negative and neutral. We used the macro F1-score as the main performance metric.

\subsection{Results of Experiments} \label{results_experiments}

We have run two series of experiments. In the first series, the training data were used to select the optimal hyperparameters for lexicon-based methods: lexicon and threshold values (see Subsection~\ref{experimental_setup}). In the second series the lexicon-based methods with selected hyperparameters were compared on test data with the fine-tuned RuBERT model.

\begin{figure}
\begin{center}
\includegraphics[width=1\textwidth]{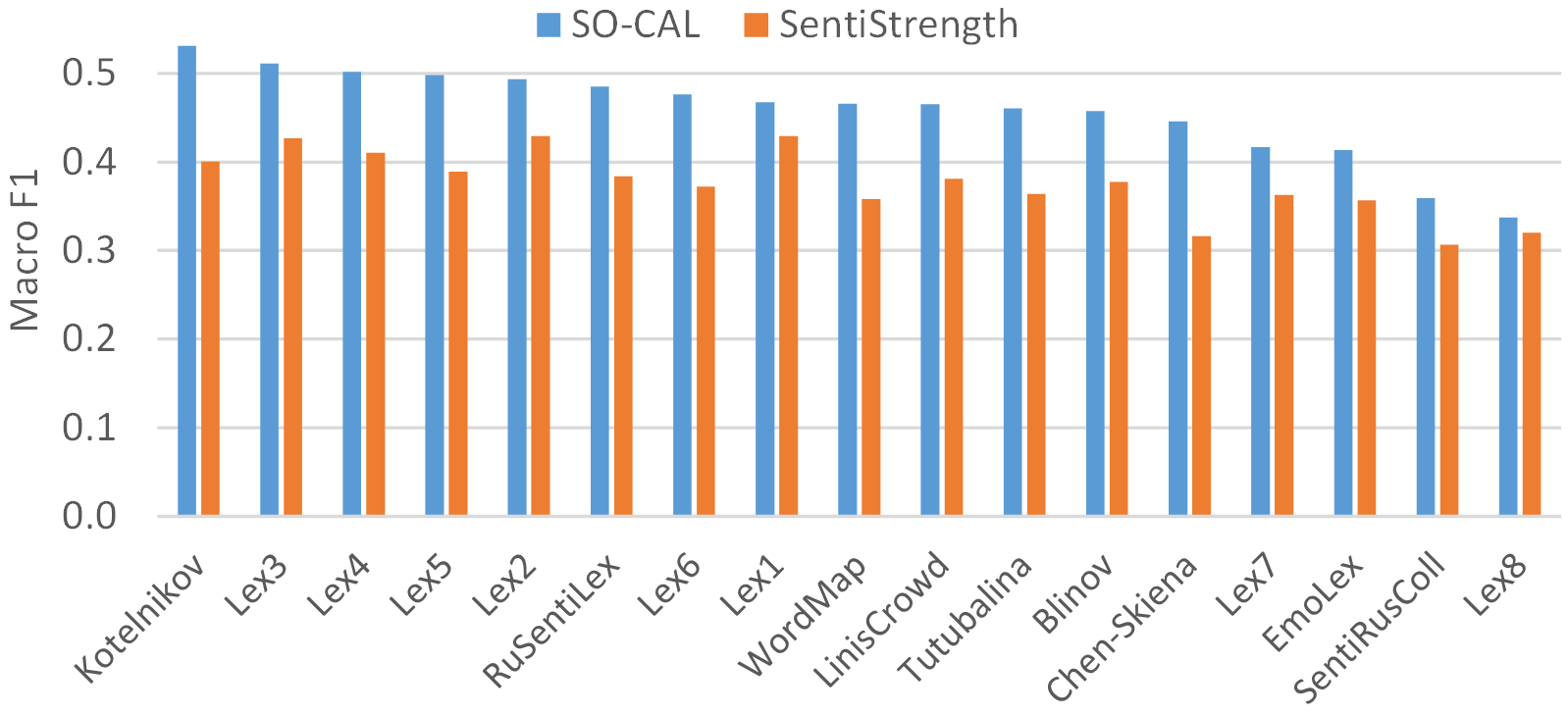}
\caption{The results of experiments on the selection of the optimal lexicon.} \label{fig1}
\end{center}
\end{figure}

The results of the first series of experiments on selecting the optimal sentiment lexicon are shown in Fig.~\ref{fig1}. Kotelnikov's lexicon has turned out to be the best lexicon for SO-CAL, \textit{Lex1} and \textit{Lex2} – for SentiStrength (\textit{Lex2} was used as the optimal lexicon, being the smaller one).

The mean values and standard deviations of the fitted thresholds for SO-CAL with Kotelnikov’s lexicon turned out to be: \(t_{1}=-1.1\pm0.95\), \(t_{2}=0.4\pm0.82\). For SentiStrength with \textit{Lex2}, coefficient \(k_{neut}=0.6\pm0.65, k=1.1\pm0.36\).

The results of the second series of experiments – the comparison of lexicon-based methods with RuBERT – are shown in Fig.~\ref{fig2}. RuBERT is superior to lexicon-based methods: on average over all the corpora for RuBERT F1-score=0.5833, for SO-CAL F1 score=0.5310, for SentiStrength F1-score=0.4290.

For 12 corpora out of 16 RuBERT outperforms both lexicon-based methods. The most significant difference was for RuSentiment (28 percentage points), Romip 2012 News (21 p.p.), tweets of SentiRuEval 2015 Banks and SentiRuEval 2016 Telecoms (20 p.p.). However, for four corpora out of 16, SO-CAL comes out on top, and for three of them the difference is quite significant: SentiRuEval 2015 Cars (32 p.p.), SentiRuEval 2015 Restaurants (29 p.p.), SemEval 2016 (25 p.p.), ROMIP 2012 Books (5 p.p.). SentiStrength, as a rule, loses to both methods, with the exception of the corpus SentiRuEval 2015 Cars.

In general, RuBERT analyzes all the corpora with short texts much better – an average number of symbols in the text less than 100 (the difference is from 13 to 28 p.p.). For medium-sized texts (700-900 symbols), the lexicon-based methods are better. For longer texts, the situation is ambiguous.

\section{Discussion}

We have compared sets of predictions on all the test corpora for all the methods. As a result, five subsets have been obtained: 1) the predictions of all methods matched, 2) the SO-CAL and SentiStrength’s predictions matched, which did not match with RuBERT; 3) the predictions from RuBERT and SO-CAL matched, which did not match with SentiStrength; 4) RuBERT and SentiStrength’s predictions matched, which did not match with SO-CAL; 5) the predictions of all the methods did not match. We have calculated the macro F1-score for each of these sets (Table~\ref{tab4}).

\begin{figure}
\begin{center}
\includegraphics[width=1\textwidth]{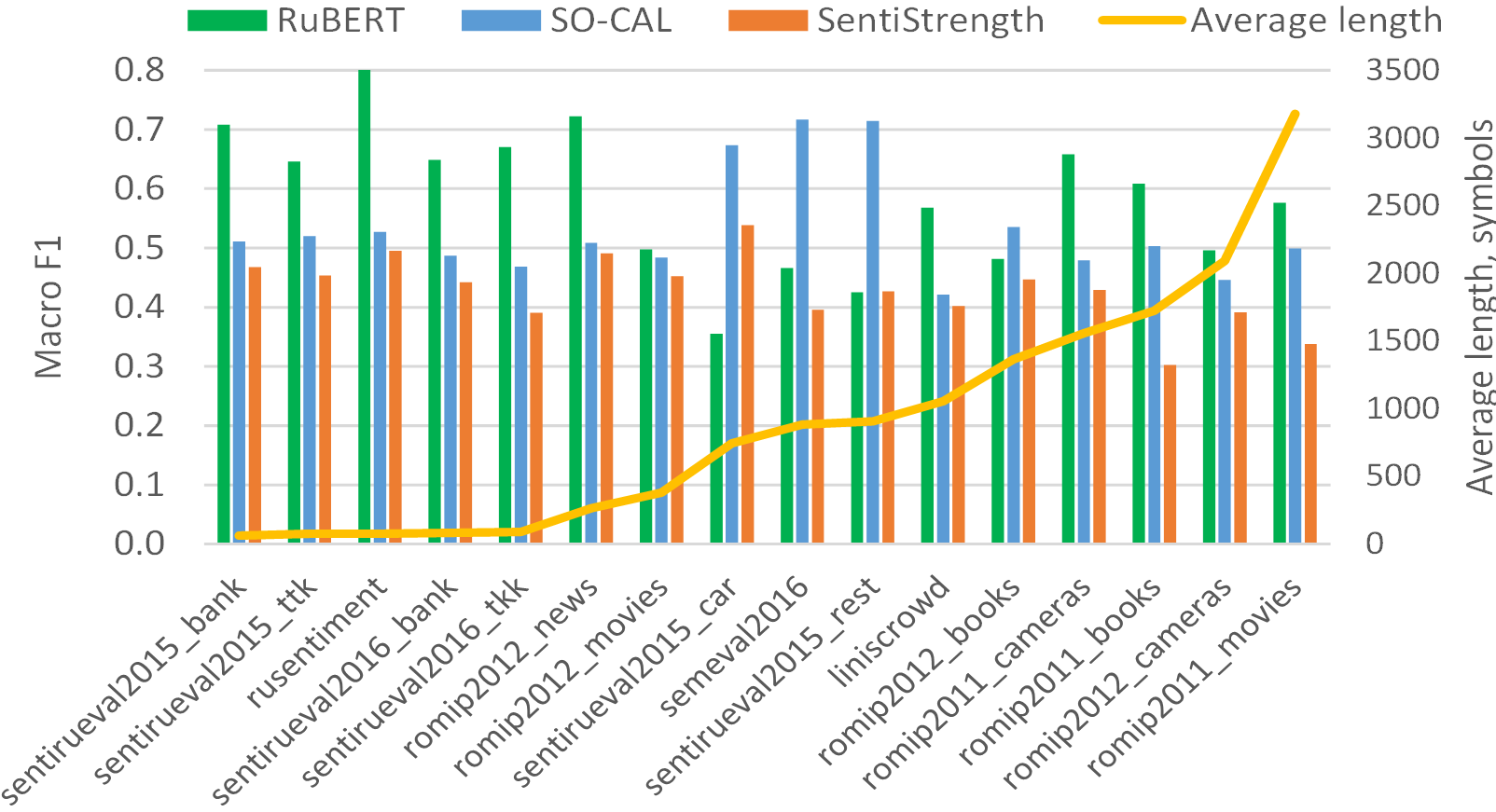}
\caption{Comparison results of lexicon-based methods and RuBERT on test corpora.} \label{fig2}
\end{center}
\end{figure} 

Table~\ref{tab4} shows that for the set of matching predictions (the first row – 38\% of the test dataset), the performance turns out to be quite high (0.8100). On average, these are shorter texts (the average length is 481 characters). For the set of matching predictions from RuBERT and SO-CAL (the third row – 18\% of the test dataset), the result is significantly higher than that of SentiStrength, which did not coincide with them – 0.7151 vs. 0.1955. For the set of identical predictions from RuBERT and SentiStrength (the fourth row – 17\% of the test sample), the difference is smaller – 0.6576 vs. 0.2347 – SentiStrength is worse than SO-CAL. Finally, for the set of the unmatched predictions (the fifth row – 6\%) RuBERT is far superior to both lexicon-based methods – 0.5617 vs. 0.1860 (SentiStrength) and 0.2074 (SO-CAL).

We analyzed in more detail the set of matching predictions of SO-CAL and SentiStrength that did not match RuBERT (the second row – 21\% of the test dataset). The results of lexicon-based methods were significantly lower than RuBERT (0.3237 vs. 0.5625). In this case, lexicon-based methods recognize positive and negative texts poorly (0.2008 and 0.3395, respectively, versus 0.5228 and 0.6654 for RuBERT). For neutral texts, the difference is not so significant – 0.4308 for lexicon-based methods vs. 0.4993 for RuBERT. If we consider the results for individual corpora, then, in general, the picture remains the same as on Fig.~\ref{fig2} with a few exceptions. The lexicon-based methods show the best results for six corpora, and not for four – two new corpora (ROMIP 2012 Cameras and Movies) were added to the previous corpora (SentiRuEval 2015 Cars and Restaurants, SemEval 2016, and ROMIP 2012 Books). Also, the ROMIP 2012 Books corpus is recognized on this set of predictions much better with lexicon-based methods than with RuBERT: 0.5000 vs. 0.0833.

\begin{table}
  \begin{center}
  \caption{Performance metrics (macro F1-score) on predictions sets.}
  \renewcommand\tabcolsep{5.0pt}
  \label{tab4}    
    \begin{tabular}{m{28mm}|p{11mm}|p{11mm}|p{11mm}|p{12mm}|p{18mm}}
    \hline
    \hline
    Set & \multicolumn{1}{c|}{RuBERT} & Senti-Strength & \multicolumn{1}{c|}{SO-CAL} &  \multicolumn{1}{c|}{Set size} & Average text length, sym.\\
    \hline
	All matched & \multicolumn{3}{c|}{0.8100} & \multicolumn{1}{c|}{14,310 (38\%)} & \multicolumn{1}{c}{481} \\ \hline
\raggedright SentiStrength \& SO‑CAL matched & \multicolumn{1}{c|}{0.5625} & \multicolumn{2}{c|}{0.3237} &  \multicolumn{1}{c|}{7,698 (21\%)} & \multicolumn{1}{c}{519} \\ \hline
\raggedright RuBERT \& SO‑CAL matched & \multicolumn{1}{c|}{0.7151} & \multicolumn{1}{c|}{0.1955} & \multicolumn{1}{c|}{0.7151} & \multicolumn{1}{c|}{6,755 (18\%)} & \multicolumn{1}{c}{603} \\ \hline
\raggedright	RuBERT \& Senti- Strength matched & \multicolumn{2}{c|}{0.6576} & \multicolumn{1}{c|}{0.2347} & \multicolumn{1}{c|}{6,289 (17\%)} & \multicolumn{1}{c}{686} \\ \hline
	All didn’t match & \multicolumn{1}{c|}{0.5617} & \multicolumn{1}{c|}{0.1860} & \multicolumn{1}{c|}{0.2074} & \multicolumn{1}{c|}{2,362 (6\%)} & \multicolumn{1}{c}{602} \\ 
	\hline
	\hline
   \end{tabular}
  \end{center}
\end{table}

We also analyzed the reasons for the incorrect predictions of the lexicon-based methods. Errors most often arise due to an insufficient size of sentiment lexicon, the absence of sentiment words in the text, an incorrect recognition of negation and irrealis, an overbalance of words of the opposite sentiment, sarcasm, and erroneous identification of domain-oriented words.

\begin{figure}
\begin{center}
\includegraphics[width=1\textwidth]{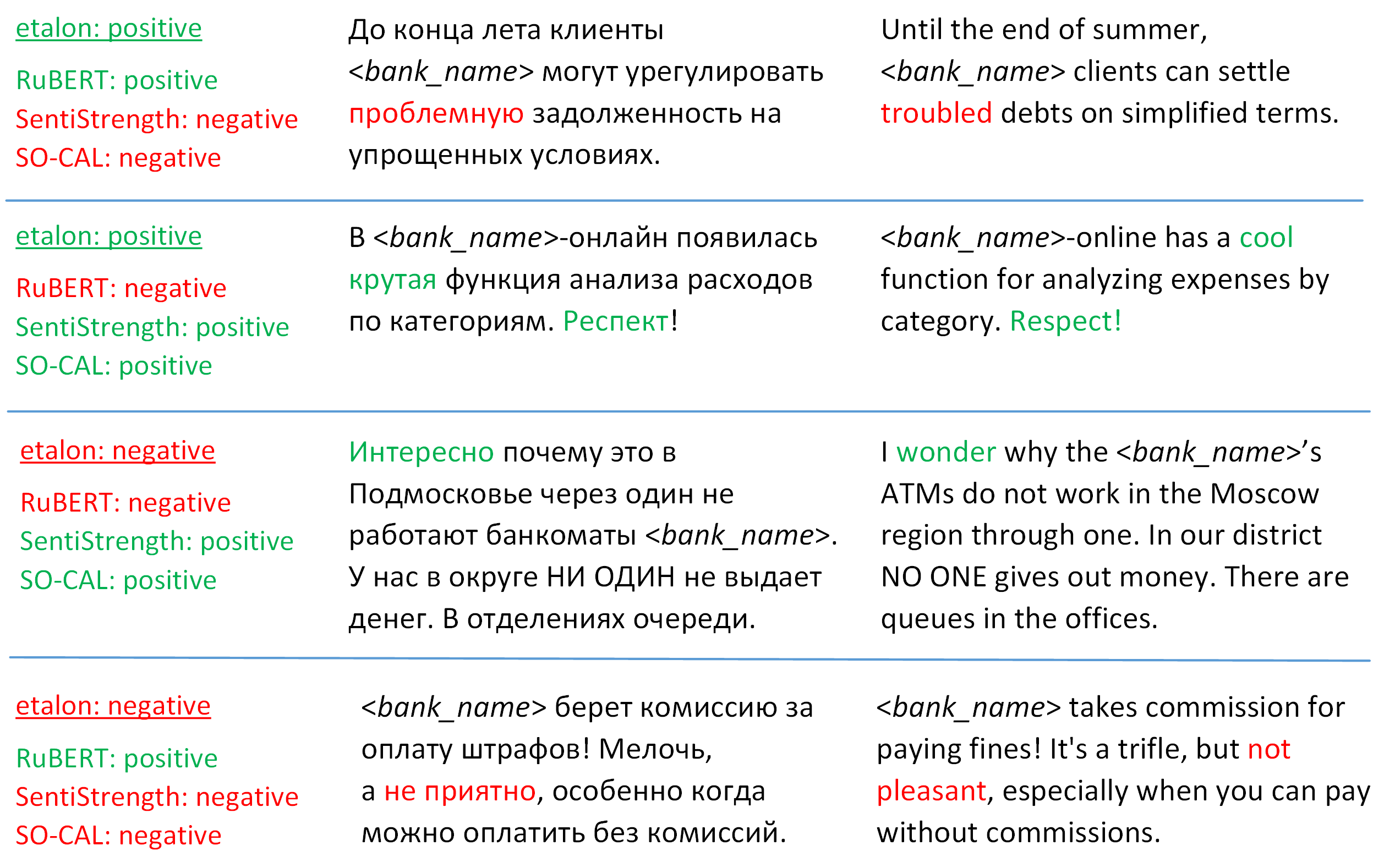}
\caption{Examples of classifiers' errors: the first and third examples are incorrectly classified by lexicon-based methods; the second and  fourth examples are misclassified by RuBERT. The first column shows real class and the answers of the methods. The second column is in Russian, the third one -- the translation into English. Sentiment words in examples are coloured: positive ones are green, negative ones are red.} \label{fig3}
\end{center}
\end{figure}

As an illustration (see Fig.~\ref{fig3}), we can give some examples that are incorrectly classified by lexicon-based methods (the first and the third examples) or RuBERT (the second and the fourth examples). In the first example lexicon-based methods didn't recognize that the phrase \textit{settle trouble debts} has positive polarity. In the third example the word \textit{wonder} led the lexicon-based methods to the wrong decision: they didn't take into account the phrase \textit{ATMs do not work}. Unfortunately, RuBERT does not have the same good interpretability as lexicon-based methods, so we can't explain why RuBERT misclassified the second and fourth examples. But as we can see from the first and third examples, RuBERT can correctly classify texts even when they have words of opposite sentiment.

\section{Conclusion}

We have compared the lexicon-based methods SO-CAL and SentiStrength with a deep neural network model RuBERT on 16 Russian-language sentiment corpora for a three-class problem of sentiment analysis. On average, RuBERT shows a higher classification performance than lexicon-based methods, exceeding SO-CAL by an average of 5 p.p. SentiStrength lags behind SO-CAL by 10 p.p. However, for four corpora out of 16 (usually with medium-length texts) SO-CAL shows better results than RuBERT. This keeps us optimistic about the lexicon-based approach in general.

In the future, we intend to modify SO-CAL in order to more accurately take into account the peculiarities of the Russian language, such as negation and irrealis. Also, a promising area of research is the development of hybrid models that combine the ability to take into account the context of deep neural networks and linguistic knowledge contained in the sentiment lexicons.

\section*{Acknowledgement}
This research is financially supported by The Russian Science Foundation, Agreement №17-71-30029 with co-financing of Bank Saint Petersburg.
	\bibliography{refs.bib}
	\bibliographystyle{splncs04}
\end{document}